\documentclass[runningheads]{llncs}

\usepackage{eccv}

\usepackage{eccvabbrv}

\usepackage{graphicx}
\usepackage{booktabs}

\usepackage[accsupp]{axessibility}  % Improves PDF readability for those with disabilities.

\usepackage{hyperref}

\usepackage{orcidlink}

\usepackage{multirow}
\usepackage{colortbl}
\usepackage{xcolor}
\usepackage{array}   %对表列和表格线的设置需要用到array宏包
\usepackage{amsmath}
\usepackage{booktabs}
\usepackage{bbding}
\usepackage{subcaption}
\usepackage{graphicx}
\usepackage{colortbl}  %彩色表格需要加载的宏包
\usepackage{xcolor}
\usepackage{array}   %对表列和表格线的设置需要用到array宏包
\usepackage{soul, color, xcolor}

\usepackage{booktabs}
\usepackage{pifont}
\usepackage{makecell}
\definecolor{lightorange}{RGB}{255, 235, 200}
\definecolor{lightblue}{RGB}{220, 235, 255}
\definecolor{mypink}{rgb}{.99,.91,.95}
\definecolor{mycyan}{cmyk}{.3,0,0,0}
\definecolor{cvprblue}{rgb}{0.21,0.49,0.74}
\usepackage{subcaption} % 若你还需要子图功能（可选）

\usepackage{etoc}    % 局部目录  \begingroup

\usepackage{bm,amsmath,amssymb}
\usepackage{wrapfig}
\usepackage{capt-of} % 提供 \captionof{figure}{...}

\begin{document}

\title{RED: Robust Event-Guided Motion Deblurring with Modality-Specific Disentanglement} 
% RED: Robust Event-Guided Motion Deblurring with Modality-Specific Disentanglement
% Robustness-Oriented Perturbation and Disentangled Representations for Event-Guided Motion Deblurring

\titlerunning{Abbreviated paper title}

\author{
Yihong Leng$^{1,2*}$,
Siming Zheng$^{2*}$,
Jinwei Chen$^{2}$,
Bo Li$^{2}$,
Jiaojiao Li$^{1}$,
Peng-Tao Jiang$^{2\dagger}$
}
\footnotetext{* Equal contribution. Work was done during intern at vivo.}
\footnotetext{$\dagger$ Corresponding author.}

% Second Author\inst{2,3}\orcidlink{1111-2222-3333-4444} \and
% Third Author\inst{3}\orcidlink{2222--3333-4444-5555}}

% % TODO FINAL: Replace with an abbreviated list of authors.

% % TODO FINAL: Replace with your institution list.
\institute{$^1$Xidian University, 
$^2$vivo Mobile Communication Co., Ltd. \\}
% Springer Heidelberg, Tiergartenstr.~17, 69121 Heidelberg, Germany

\maketitle

\begin{abstract}
Event-guided motion deblurring reconstructs sharp images using the high-temporal-resolution motion cues from event cameras.
However, in real capture, thresholding-induced event under-reporting causes missing and fragmented motion cues, under which existing methods often degrade in performance due to two limitations: i) assumptions of dense and stable events, and ii) modality-indiscriminate extraction and fusion that fail to separate useful motion cues from disrupted events, allowing them to contaminate cross-modal representations.
In this paper, we first introduce a Robustness-Oriented Perturbation Strategy (RPS) that mimics various trigger thresholds of dynamic vision sensors, exposing our model to diverse under-reporting patterns and thereby improving robustness under unknown conditions.
Built upon this setting, we propose RED, a Robust Event-guided Deblurring network, following the principle of disentangle first and then fuse selectively. 
Specifically, the Modality-specific Representation Mechanism disentangles the inputs into image-semantic, event-motion, and cross-modal representations, capturing appearance, motion, and complementary interactions, respectively. 
With the reliable disentangled features, we selectively fuse modalities to enhance motion-sensitive areas in blurry images and enrich under-reported events with semantic context. 
Extensive experiments on synthetic and real-world datasets demonstrate RED consistently achieves state-of-the-art performance in terms of both accuracy and robustness.
\keywords{Event camera \and Image deblur \and Robustness}
\end{abstract}

\vspace*{-26pt}
\section{Introduction}
Motion blur is a pervasive degradation in dynamic visual scenes, typically caused by rapid object motion or unintended camera shake during exposure.
Image deblurring aims to reconstruct sharp textures and structures from a blurred observation, with approaches spanning handcrafted priors \cite{levin2009understanding, krishnan2011blind, bahat2017non} to modern CNN- and Transformer-based models \cite{nah2017deep, tao2018scale, dong2023multi, kong2023efficient, liang2024image, wang2024correlation}.
Yet under severe motion blur, critical structural and temporal details are heavily corrupted, and the performance of these methods degrades markedly.

Event cameras, inspired by biological vision systems, have emerged as a promising alternative to conventional sensors in high-speed scenes \cite{gallego2020event}.
They emit asynchronous streams of events that encode motion with high temporal precision, making them naturally well suited for motion deblurring. 
Early methods \cite{pan2019bringing, zhang2022unifying} relied on physical models to establish deterministic mappings between blurred images and events, whereas recent studies have adopted deep learning-based solutions that enable more expressive and powerful representations.
In particular, cross-modal fusion strategies such as event-image attention mechanisms \cite{sun2022event}, spatio-temporal attention modules \cite{yang2024motion}, and frequency-aware interactions \cite{kim2024frequency} have been introduced to bridge events and blurry images. 
Furthermore, feature interaction strategies such as unidirectional event-guided fusion \cite{sun2023event}, bidirectional modality alignment \cite{chen2024motion}, and asymmetric bidirectional integration \cite{yang2025asymmetric} further explore the synergistic potential of multi-modal features.

\begin{figure}[t]
    \centering
    \begin{minipage}[t]{0.38\textwidth}
        \centering
        \includegraphics[width=\linewidth]{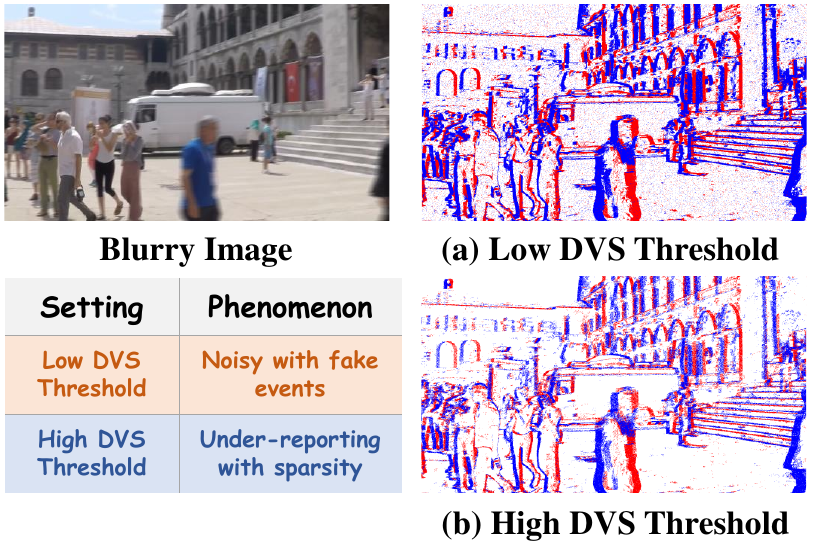}
        \caption{Threshold-driven under-reporting: weak motions that fall below the trigger condition are not reported.}
        \label{fig1}
    \end{minipage}\hfill
    \begin{minipage}[t]{0.60\textwidth}
        \centering
        \includegraphics[width=\linewidth]{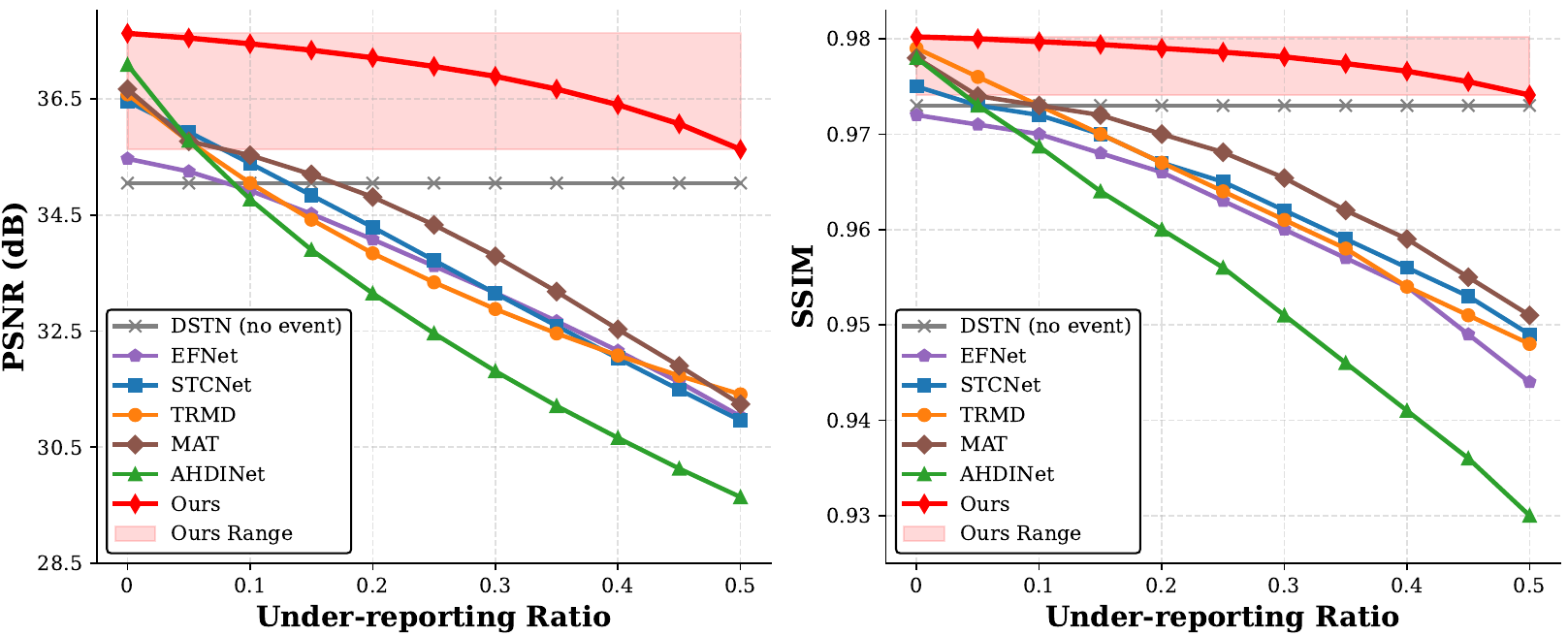}
        \caption{ As the under-reporting ratio increases, the performances of existing event-based deblurring methods degrade sharply and fall below the image-only method DSTN.}
        \label{fig2}
    \end{minipage}
    \vspace*{-20pt}
\end{figure}

In practice, event cameras are often operated with higher trigger thresholds of dynamic vision sensors (DVS) to suppress unstable firings and obtain more reliable event streams. However, this inevitably leads to threshold-driven event under-reporting, where events from weak motions or low-contrast edges fall below the trigger condition and are therefore not reported (Fig.~\ref{fig1}). 
As the under-reporting ratio increases, existing event-guided deblurring methods degrade sharply in Fig.~\ref{fig2} and can even underperform the image-only baseline\cite{Pan_2023_CVPR}. 
This contradicts the common assumption that events monotonically contribute to deblurring: under pronounced under-reporting, the marginal utility of events becomes non-positive and can even be detrimental.
We attribute this degradation to the fact that i) current pipelines are typically developed under the assumption of dense and stable events and ii) often adopt modality-indiscriminate feature extraction and naive fusion, which fail to isolate useful motion cues from disrupted events and instead contaminate cross-modal representations.
% This motivates us a ‘disentangle-first, fuse-selectively’ design to prevent fragmented events from dominating fusion while still exploiting their remaining motion cues.

To tackle the above challenges, we present RED, a Robust Event-guided Deblurring network built on modality-specific disentanglement.
First, we introduce a Robustness-Oriented Perturbation Strategy (RPS) that mimics realistic threshold-induced under-reporting by varying DVS trigger thresholds, exposing the network to diverse event dropouts and improving robustness under unknown capture conditions. 
Built upon this setting, RED follows the principle of `disentangle first and then fuse selectively'.
In detail, the Modality-specific Representation Mechanism (MRM) is designed to disentangle semantic, motion, and cross-modality features. 
By separating semantic information from images and motion details from events before fusion, MRM prevents corrupted event characteristics from overwhelming image semantics and allows our RED to retain useful motion priors. 
With these reliable disentangled features, two modules are designed to achieve coadjutant interactions: 
the Motion Saliency Enhancer Module (MSEM) that transfers motion-sensitive priors to enhance spatial details easily lost in blur, and
the Event Semantic Engraver Module (ESEM) that delineates semantic characteristics from images into deeper event encoding, mitigating the semantic deficiency caused by sparse events.
In this way, RED achieves reliable deblurring through robust event adaptability, disentangled representation, and selective interaction.
In summary, our main contributions are as follows:

\begin{enumerate}
    \item{
    We propose RED for robust event-guided motion deblurring. 
    Extensive experiments demonstrate our RED outperforms existing methods in both deblurring quality and robustness on synthetic and real-world datasets.
    }
    \item{ 
    RPS exposes our RED to diverse under-reporting events, enhancing the robustness and adaptability to real-world conditions.
    }
    \item{
    % 这里要加可视化来进一步说明所起的作用，来进一步说明其点
    {MRM first factorizes feature space into semantic and temporal dimensions for modality-specific disentanglement, and then robust motion-sensitive priors and compensatory semantic context are interacted by MSEM and ESEM.}
    }
\end{enumerate}

\vspace*{-20pt}
\section{Related Work}
\vspace{-6pt}
\subsection{Image Deblurring}
% task methods
Image deblurring seeks to recover a clear image from its motion-blurred counterpart, a degradation typically caused by camera shake or dynamic object motion. 
Classical methods typically follow a two-step pipeline: estimating the blur kernel and performing non-blind deconvolution \cite{fergus2006removing, cho2009fast}. 
To constrain this ill-posed problem, various image priors, such as Total Variation \cite{chan1998total}, sparsity \cite{xu2013unnatural},  patch recurrence \cite{michaeli2014blind}, and color statistics\cite{pan2016blind} have been explored. 
While effective in controlled settings, these approaches struggle with spatially-varying blur and rely heavily on accurate kernel modeling \cite{tran2021explore, hyun2013dynamic}, often leading to unstable performance and high computational cost.
With the advent of deep learning, CNN-based architectures have demonstrated notable advances by directly learning the blur-to-sharp mapping. 
Multi-scale feature fusion \cite{nah2017deep, tao2018scale, dong2023multi}, multi-stage refinement \cite{chen2021hinet, zamir2021multi, yang2022progressive, liu2024transformer}, and coarse-to-fine strategies \cite{cho2021rethinking, zheng2022progressive} have been extensively explored.
More recently, attention-based models \cite{tsai2022stripformer, zamir2022restormer, kong2023efficient, mao2024loformer} have achieved state-of-the-art performance by capturing long-range dependencies and preserving fine-grained structures. 
Despite these advancements, existing models still struggle under severe or complex blur conditions.

\vspace{-10pt}
\subsection{Event-based Motion Deblurring}
\vspace{-6pt}
Event cameras offer low-latency sensing by asynchronously recording per-pixel brightness changes, making them well-suited for high-speed motion deblurring.
Early works such as EDI \cite{pan2019bringing} attempted to explicitly model the physical relationship, while recent methods have leveraged deep learning to construct more robust and expressive representations.
A straightforward approach concatenated event with RGB images to assist image deblurring \cite{jiang2020learning}, but such naive integration failed to capture fine-grained interactions across modalities. 
D2Nets \cite{shang2021bringing} designed a learnable weight matrix to embed event priors into arbitrary image deblurring networks. 
Further advancing event representation, EFNet \cite{sun2022event} introduced a symmetric cumulative event encoding combined with a cross-attention mechanism to enhance inter-modal feature interaction.
To better exploit modality complementarity and suppress redundancy, STCNet \cite{yang2024motion} designed a cross-modal co-attention module that dynamically fuses spatial features from both streams.
FEVD \cite{kim2024frequency} presented a frequency-domain filtering approach that captures global cross-modal dependencies via spatial frequency interactions. 
Most recently, AHDINet \cite{yang2025asymmetric} argued that symmetric or unidirectional fusion strategies \cite{sun2023event, chen2024motion} may lead to either insufficient or redundant interactions. They instead proposed an asymmetric bidirectional integration framework to more effectively leverage complementary information.

Despite these advances, most existing methods implicitly assume that event streams are complete and reliable. 
In practice, however, events are relatively under-reporting due to the thresholding mechanism of DVS. 
With an ignorance of the challenging and disrupted events, shared feature extractors or generic fusion strategies in current pipelines fail to effectively disentangle semantic context from images and motion evidence from practical events, leading to fragile cross-modal representations and reduced robustness.

\vspace{-10pt}
\section{Methods}

\vspace{-6pt}
\subsection{Robustness-Oriented Perturbation Strategy (RPS)} 
\vspace{-6pt}
Event cameras generate an event when the logarithmic intensity change at a pixel exceeds a 
contrast threshold $\theta \in \mathbb{R}_{+}$.    
Formally, let the log-intensity increment at pixel coordinates $(x,y)$ and time $t$ be
\vspace*{-4pt}
\begin{equation}
\Delta \ell(x,y,t) = \log I_t(x,y) - \log I_{t-\Delta t}(x,y),
\end{equation}
where $I_t(x,y)$ denotes the intensity at time $t$, and $\Delta t$ is the time interval.  
We decompose $\Delta \ell$ into a signal and a noise component, which originates from multiple physical sources~\cite{jiang2024edformer}, such as photon shot noise and thermal or electronic fluctuations, which can be modeled as Poisson and Gaussian process:
\vspace*{-4pt}
\begin{equation}
\Delta \ell = S + N = N = N_p + N_g,
\qquad N_p \sim \mathrm{Poisson}(\lambda), 
\qquad N_g \sim \mathcal{N}(0,\sigma_n^2),
\end{equation}
where $S$ corresponds to true motion signals, $N$ accounts for the unified noise, $N_p$ accounts for photon arrivals with rate $\lambda$, and $N_g$ represents circuit-induced perturbations with variance $\sigma_n^2$.   

An event is triggered when $|\,\Delta \ell\,| \ge \theta, p = \mathrm{sign}(\Delta \ell)$.
However, when $|N_p + N_g| \ge \theta$ or $S$ is weak, an event is incorrectly triggered.
The corresponding false positive rate (FPR) with $\mathbb{P}(\cdot)$ as the probability of an event is:\
\vspace*{-4pt}
\begin{equation}
\mathrm{FPR}(\theta) = \mathbb{P}\big(|N_p+N_g| \ge \theta\big),
\end{equation}
which decreases as $\theta$ increases. 
When true motion $S$ is present, the true positive rate (TPR) is: $\mathrm{TPR}(\theta \mid S) 
= \mathbb{P}\big(|S+N_p+N_g| \ge \theta \;\mid\; S\big).$

Although larger thresholds reduce $\mathrm{FPR}$, $\mathrm{TPR}$ is decreased by discarding weak but informative events.  
We quantify this by the under-reporting ratio (UR):
{
\setlength{\abovedisplayskip}{4pt}
\setlength{\belowdisplayskip}{2pt}
\setlength{\abovedisplayshortskip}{4pt}
\setlength{\belowdisplayshortskip}{2pt}
\begin{equation}
\mathrm{UR}(\theta \mid S) = 1 - \mathrm{TPR}(\theta \mid S),
\end{equation}
}
whose expectation over the distribution of $S$ defines the overall under-reporting:
{
\setlength{\abovedisplayskip}{2pt}
\setlength{\belowdisplayskip}{2pt}
\setlength{\abovedisplayshortskip}{2pt}
\setlength{\belowdisplayshortskip}{2pt}
\begin{equation}
\mathrm{UR}(\theta) = \mathbb{E}_S \big[\, \mathrm{UR}(\theta \mid S) \,\big].
\end{equation}
}

\begin{figure}[!t]
	\centering
	\includegraphics[width=0.98\textwidth]{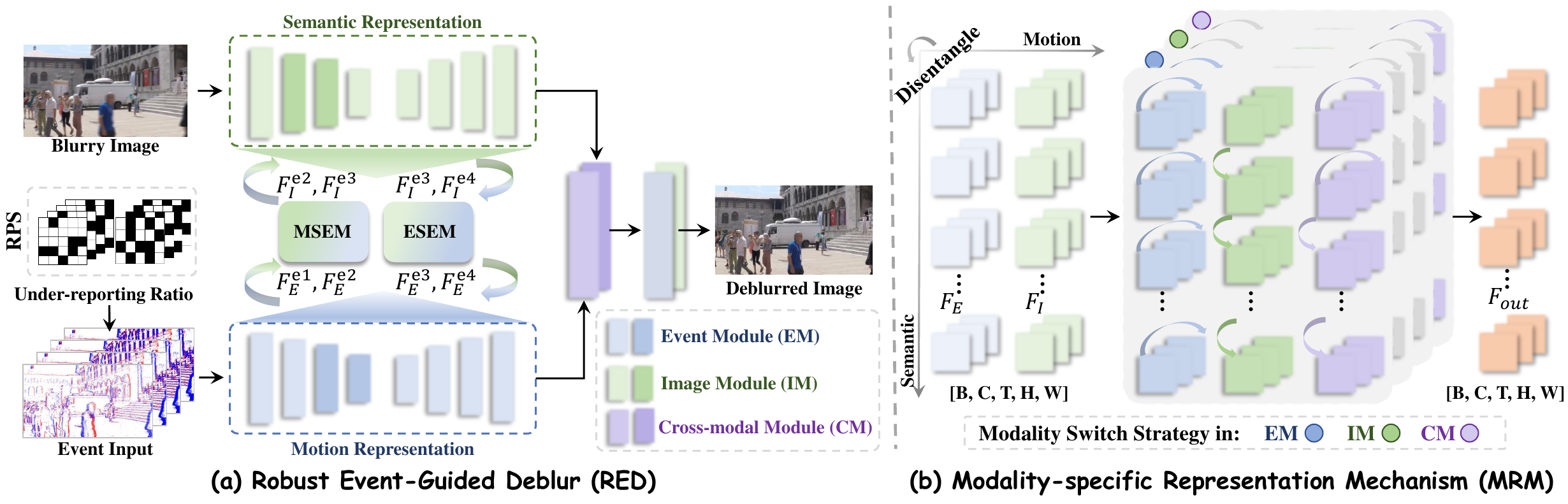}%
    \vspace{-4pt}
	\caption{ \textbf{Overview of our RED. } 
    In detail, a Robustness-Oriented Perturbation Strategy (RPS) is implemented to event input, and MRM is designed to disentangle modality-specific features with individually semantic reasoning in image module, motion-wise representation in motion module, and cross-modality fusion in cross-modal module. Furthermore, Motion Saliency Enhancer Module (MSEM) is designed to excavate motion-sensitive priors to image branch and Event Semantic Engraver Module (ESEM) is presented to compensate events with global semantic understanding.
    }
\label{framework}
\vspace*{-14pt}
\end{figure}

In summary, $\theta$ acts as a global control knob:
$\theta \uparrow \;\Rightarrow\; \mathrm{FPR} \downarrow, \; \mathrm{UR} \uparrow.$
This motivates our RPS, which explicitly reformulates event acquisition as a probabilistic triggering process.  
At pixel $(x,y)$ and time $t$, whether an event survives is determined by the survival probability:
{
\setlength{\abovedisplayskip}{2pt}
\setlength{\belowdisplayskip}{2pt}
\setlength{\abovedisplayshortskip}{2pt}
\setlength{\belowdisplayshortskip}{2pt}
\begin{equation}
\pi_t(x,y) = \mathbb{P}\big(|S+N_p+N_g| \ge \theta\big),
\end{equation}
}
which encodes the likelihood that the log-intensity increment surpasses the contrast threshold.  
Thus, the problem naturally reduces to a binary decision, an event either survives or is suppressed.  
To emulate this mechanism during training, we adopt the voxel grid representation $\mathbf{D}\in\mathbb{R}^{T\times H\times W}$ of events ~\cite{sun2022event}, where $T$ denotes the number of temporal bins.  
At each slice $\tau \in \{1,\dots,T\}$, we stochastically thin the events according to their survival probability:
{
\setlength{\abovedisplayskip}{2pt}
\setlength{\belowdisplayskip}{2pt}
\setlength{\abovedisplayshortskip}{2pt}
\setlength{\belowdisplayshortskip}{2pt}
\begin{equation}
\widetilde{\mathbf{D}}_{\tau} = \mathbf{D}_{\tau} \odot \boldsymbol{\rho}_\tau, 
\qquad \boldsymbol{\rho}_\tau(x,y) \sim \mathrm{Bernoulli}\!\big(\pi_{\tau}(x,y)\big),
\label{bernolli}
\end{equation}
}
where $\odot$ denotes element-wise multiplication.  
During training, we further use $\mathrm{UR}$ to control perturbation strength.     
In detail, for each iteration we sample
{
\setlength{\abovedisplayskip}{2pt}
\setlength{\belowdisplayskip}{2pt}
\setlength{\abovedisplayshortskip}{2pt}
\setlength{\belowdisplayshortskip}{2pt}
\begin{equation}
\alpha \sim \mathcal{U}(\alpha_{\min}, \alpha_{\max}), 
\qquad \frac{1}{HW}\sum_{x,y}\pi_\tau(x,y) \approx 1-\alpha, \quad \forall \tau,
\label{ur}
\end{equation}
}
and draw $\boldsymbol{\rho}_\tau$ independently across temporal bins to mimic variability in thresholds and circuitry.  
This generates a continuum of training regimes from mild to severe under-reporting, faithfully reflecting the physical mechanism of threshold-driven event triggering.  
As verified in Figure~\ref{fig_ablation_rps}, our architecture-agnostic and parameter-free RPS substantially improves the robustness of the network.

\vspace{-12pt}
\subsection{Modality-specific Representation Mechanism (MRM)}
Blurry images primarily encode high-level semantic context, whereas under-reporting events provide complementary motion priors but lack complete semantic information.  
When both modalities are indiscriminatively extracted or naively processed together, the lack of disentangling can cause semantic and motion features to mix, reducing the clarity of the extracted representations. 
This motivates us a ‘disentangle-first, fuse-selectively’ design to prevent fragmented events from dominating fusion while still exploiting their remaining motion cues.

\noindent{\textbf{Semantic-wise and Motion-wise Attention.}}
Given the intermediate features 
$\mathbf{F}^{ei}_{\mathrm{I}} $ and $\mathbf{F}^{ei}_{\mathrm{E}} \in \mathbb{R}^{B \times N \times H \times W}$ from the image and event branches, where $B$ denotes the batch size and $N$ refers to the original mixed space.
To emphasize semantic understanding, we begin by disentangling the feature space. Using a $1{\times}1$ convolution followed by a depthwise $3{\times}3$ convolution, we obtain the query, key, and value tensors for semantic modeling:
{
\setlength{\abovedisplayskip}{2pt}
\setlength{\belowdisplayskip}{2pt}
\setlength{\abovedisplayshortskip}{2pt}
\setlength{\belowdisplayshortskip}{2pt}
\begin{equation}
\mathbf{Q}_{\text{sem}}, \mathbf{K}_{\text{sem}}, \mathbf{V}_{\text{sem}} \in \mathbb{R}^{B \times L \times C_{L} \times (T  H  W)},
\end{equation}
}
where $N = C \times T$ represents the split between the channel and temporal dimensions, and $C = C_L \times L$ denotes the multi-head splitting, with $L$ being the number of heads.  
For motion-specific modeling, tokens are reshaped as 
$\mathbf{Q}_{\text{mot}},\mathbf{K}_{\text{mot}},\mathbf{V}_{\text{mot}} \in 
\mathbb{R}^{B \times L \times T_{L} \times (C H W)}$, 
where $T = L \times T_{L}$. 
Here, the splitting focuses on temporal indices, ensuring the attention explicitly captures motion continuity across time.
Thus, the corresponding semantic correlation and motion dependency are formulated as
{
\setlength{\abovedisplayskip}{2pt}
\setlength{\belowdisplayskip}{2pt}
\setlength{\abovedisplayshortskip}{2pt}
\setlength{\belowdisplayshortskip}{2pt}
\begin{equation}
\mathbf{A}_{\text{sem}} = \mathrm{Softmax}(\mathbf{Q}_{\text{sem}} \mathbf{K}_{\text{sem}}^{\top}), 
\quad
\mathbf{A}_{\text{mot}} = \mathrm{Softmax}(\mathbf{Q}_{\text{mot}} \mathbf{K}_{\text{mot}}^{\top}).
\end{equation}
}
Then, the updated features are then computed as
{
\setlength{\abovedisplayskip}{2pt}
\setlength{\belowdisplayskip}{2pt}
\setlength{\abovedisplayshortskip}{2pt}
\setlength{\belowdisplayshortskip}{2pt}
\begin{equation}
\mathbf{F}^{ei}_{\mathrm{I}} = \mathbf{V}_{\text{sem}} \odot \mathbf{A}_{\text{sem}}, 
\quad
\mathbf{F}^{ei}_{\mathrm{E}} = \mathbf{V}_{\text{mot}} \odot \mathbf{A}_{\text{mot}}.
\end{equation}
}
In this way, semantic characteristics are enhanced in the image branch, while temporal motion dependencies are emphasized in the event branch.

\begin{figure}[t]
    \centering
    \begin{minipage}[t]{0.38\textwidth}
        \centering
        \includegraphics[width=\linewidth]{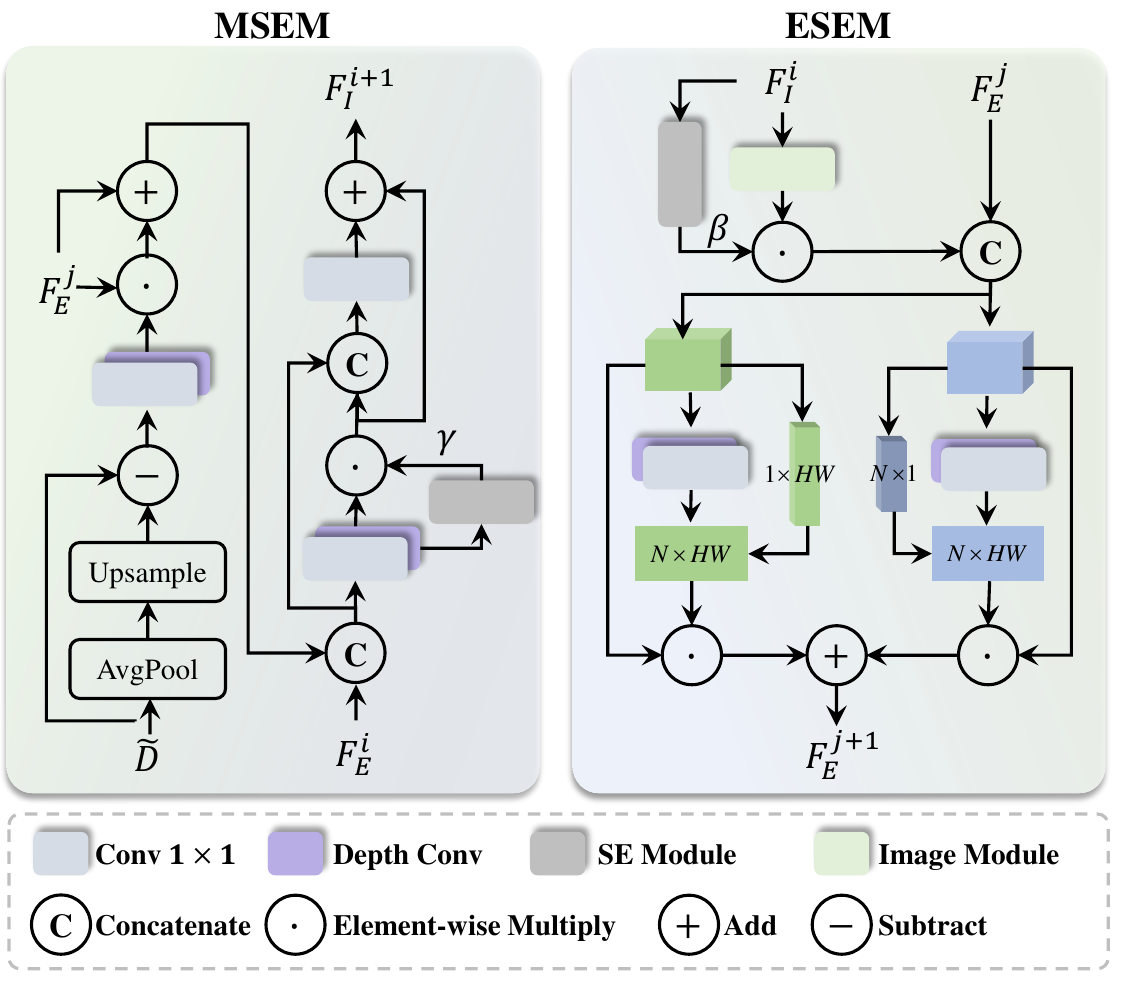}
        \caption{ Framework of our proposed Motion Saliency Enhancer Module (MSEM) and Event Semantic Engraver Module (ESEM).}
        \label{fig:msem_esem}
    \end{minipage}
    \begin{minipage}[t]{0.60\textwidth}
        \centering
        \includegraphics[width=\linewidth]{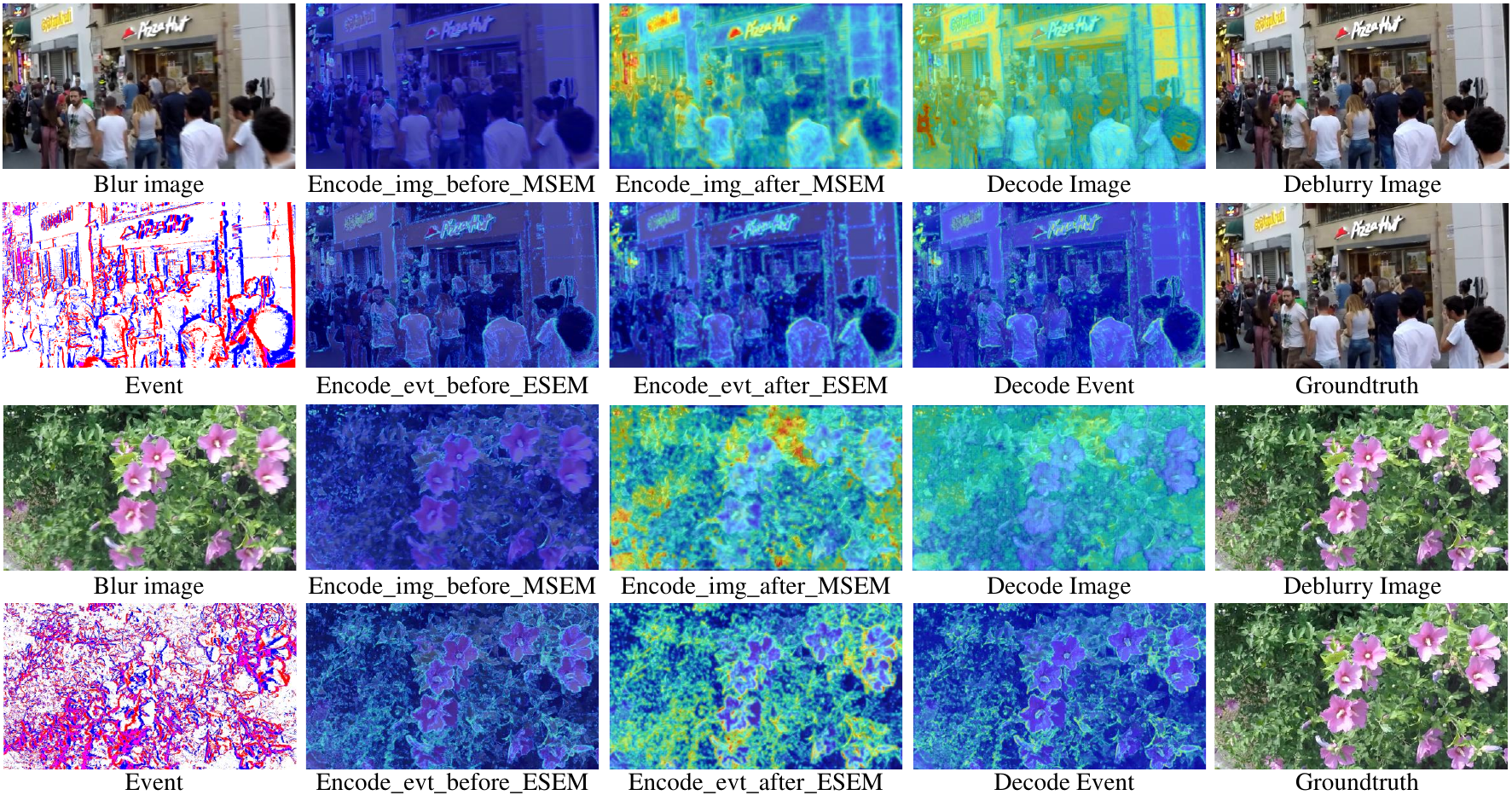}
        \caption{Visualizations of the activation maps, including: 1) image-encoder features before and after MSEM, 2) image-branch decoder features, 3) event-encoder features before and after ESEM, and 4) event-branch decoder features.}
        \label{vis_activation_map}
    \end{minipage}\hfill
    \vspace*{-18pt}
\end{figure}

% \begin{figure}[!t]
%   \centering
%   \includegraphics[width=0.95\linewidth]{Fig/Reb/Sup_Heatmap_scene56.pdf}
%   \caption{\textbf{ Visualizations.} In detail, we visualize the activation maps, including: 1) image-encoder features before and after MSEM, 2) image-branch decoder features, 3) event-encoder features before and after ESEM, and 4) event-branch decoder features.  }
%   \label{vis_activation_map}
%   \vspace{-8pt} % Best viewed on a screen and zoomed in.
% \end{figure}

\noindent{\textbf{Cross-Modality Attention.}}
With the former modality-specific representation,  a cross-modality interaction are designed to broadcast semantic understanding and motion changes.
Considering the event modality lacks global semantic structure under high under-reporting ratios, whereas the image modality still provides stable and complete semantic priors.
On the one hand, when transferring semantic cues from images to events, both the query and key are taken from the image modality to construct a reliable semantic attention map. 
In detail, query and key are derived from $\mathbf{F}^{ei}_{\mathrm{I}}$, 
while value is taken from $\mathbf{F}^{ei}_{\mathrm{E}}$.
Here, spatio-temporal tokens are chosen, thus features from images attributes to delineating abundant semantic understanding.
{
\setlength{\abovedisplayskip}{2pt}
\setlength{\belowdisplayskip}{2pt}
\setlength{\abovedisplayshortskip}{2pt}
\setlength{\belowdisplayshortskip}{2pt}
\begin{equation}
\mathbf{A}_{I \rightarrow E} = \mathrm{Softmax}\!\left(\mathbf{Q}_{I}\mathbf{K}_{I}^{\top}\right), 
\qquad
\widetilde{\mathbf{F}}_{\mathrm{E}} = \mathbf{V}_{E} \odot \mathbf{A}_{I \rightarrow E}.
\end{equation}
}
On the other hand, when transferring motion details from events to images, both the query and key are taken from the event modality to construct a reliable motion dependency. 
In detail, query and key are derived from $\mathbf{F}^{ei}_{\mathrm{E}}$, 
while value comes from $\mathbf{F}^{ei}_{\mathrm{I}}$.  
Here, spatio-channel tokens are chosen, thus events guides the images to restore detailed structural distributions.
{
\setlength{\abovedisplayskip}{2pt}
\setlength{\belowdisplayskip}{2pt}
\setlength{\abovedisplayshortskip}{2pt}
\setlength{\belowdisplayshortskip}{2pt}
\begin{equation}
\mathbf{A}_{E \rightarrow I} = \mathrm{Softmax}\!\left(\mathbf{Q}_{E}\mathbf{K}_{E}^{\top}\right), 
\qquad
\widetilde{\mathbf{F}}_{\mathrm{I}} = \mathbf{V}_{I} \odot \mathbf{A}_{E \rightarrow I}.
\end{equation}
}
Finally, two-mode outputs are concatenated to obtain the fused output $\mathbf{F}_{\text{out}}$.
In conclusion, MRM switches among semantic-wise, motion-wise, and cross-modality attentions, seperately serving as the core mechanism of image, event, and cross-modal modules.
By this way, MRM embody our RED with modality-specific representations and effective cross-modality fusion.

\vspace*{-10pt}
\subsection{MSEM and ESEM}
\vspace*{-2pt}

In our whole RED framework in Figure \ref{framework}, Motion Saliency Enhancer Module (MSEM) and  Event Semantic Engraver Module (ESEM) are designed to achieve coadjutant interactions.
As illustrated in Figure \ref{fig:msem_esem}, MSEM aims to highlight motion-sensitive structures from events and inject them into the image branch.  

Given the perturbed event input $\widetilde{\mathbf{D}}$, we first derive high-frequency components 
$\mathbf{S} \in \mathbb{R}^{B \times N \times H \times W}$ via a downsample-then-subtract operation followed by depthwise convolutions.
The motion-enhanced event feature is initialized as:
{
\setlength{\abovedisplayskip}{2pt}
\setlength{\belowdisplayskip}{2pt}
\setlength{\abovedisplayshortskip}{2pt}
\setlength{\belowdisplayshortskip}{2pt}
\begin{equation}
\widehat{\mathbf{F}}^{(1)}_{\mathrm{E}} = \mathbf{F}^{(j)}_{\mathrm{E}} \odot \mathbf{S} + \mathbf{F}^{(j)}_{\mathrm{E}},
\end{equation}
}
where $\odot$ denotes element-wise multiplication.
To transfer motion priors into the image branch, 
$\widehat{\mathbf{F}}^{(1)}_{\mathrm{E}}$ is concatenated with $\mathbf{F}^{(i)}_{\mathrm{I}}$ 
and processed by a $1{\times}1$ convolution followed by two depthwise convolutions, 
yielding a coarse fused representation $\mathbf{F}_{\text{mix}}$.  
From $\mathbf{F}_{\text{mix}}$, we compute a motion-aware attention map 
$\bm{\gamma} \in \mathbb{R}^{B \times N \times H \times W}$. 
Using this map, the event feature is refined as:
$\widehat{\mathbf{F}}^{(2)}_{\mathrm{E}} = \bm{\gamma} \odot \widehat{\mathbf{F}}^{(1)}_{\mathrm{E}}.$
Finally, $\mathbf{F}_{\text{mix}}$ is concatenated with $\widehat{\mathbf{F}}^{(2)}_{\mathrm{E}}$ 
to produce a fine-grained motion-enhanced feature $\mathbf{F}_{\mathrm{E2I}}$, 
which is then propagated to update the next-stage image representation $\mathbf{F}^{(i+1)}_{\mathrm{I}}$.

Complementary to MSEM, ESEM engraves high-level semantic representations from the image branch into the motion branch.  
Given $\mathbf{F}^{(i)}_{\mathrm{I}}$, 
we first extract a latent spatial feature $\widehat{\mathbf{F}}^{(i)}_{\mathrm{I}}$ 
through an image encoder.  
A channel-wise attention then generates correlation weights $\bm{\beta}$, and 
the semantic embedding is formulated as: 
$\mathbf{F}^{(i)}_{\mathrm{sem}} = \bm{\beta} \odot \widehat{\mathbf{F}}^{(i)}_{\mathrm{I}}.$
To enhance event features with this semantic prior, 
$\mathbf{F}^{(i)}_{\mathrm{sem}}$ is fused with $\mathbf{F}^{(i)}_{\mathrm{E}}$ 
to form $\widehat{\mathbf{F}}_{\text{mix}}$.  
We split $\widehat{\mathbf{F}}_{\text{mix}}$ along the channel dimension into 
$\widehat{\mathbf{F}}^{\mathrm{E}}_{\text{mix}}$ and 
$\widehat{\mathbf{F}}^{\mathrm{sem}}_{\text{mix}}$, 
which are individually processed by a dual-branch attention module:  
temporal attention on $\widehat{\mathbf{F}}^{\mathrm{E}}_{\text{mix}}$ produces 
an event-aware output $\overline{\mathbf{F}}^{(i)}_{\mathrm{E}}$, 
while spatial attention on $\widehat{\mathbf{F}}^{\mathrm{sem}}_{\text{mix}}$ 
produces a spatially modulated feature $\overline{\mathbf{F}}^{(i)}_{\mathrm{sem}}$.  
The two outputs are concatenated to yield the final semantically enriched 
event feature $\mathbf{F}_{\mathrm{I2E}}$, 
which is forwarded to the next-stage event representation 
$\mathbf{F}^{(i+1)}_{\mathrm{E}}$.

As illustrated in Fig.~\ref{vis_activation_map}:
i) The image branch preserves its semantic structure after MSEM, but exhibits enhanced responses exactly at event-indicated motion areas. 
ii) The event branch receives semantic reinforcement from the image branch after ESEM, producing motion representations with more complete contours and improved structural continuity. 

\begin{figure}[!t]
  \centering
  \includegraphics[width=0.95\linewidth]{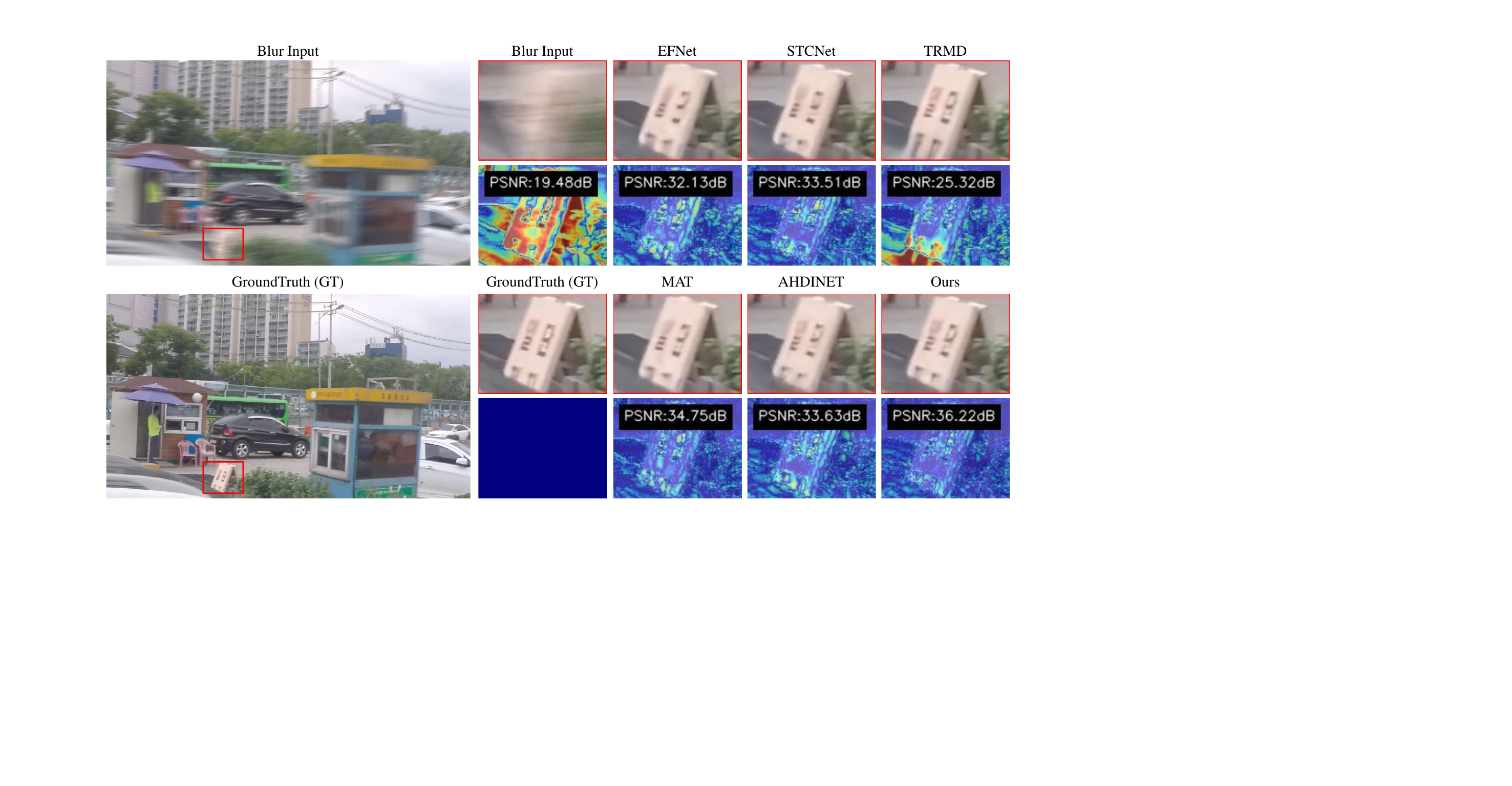}
  \vspace{-4pt}
  \caption{\textbf{Visualization in GoPro dataset.} Besides, error maps are drawn to illustrate a comprehensive comparison of both localization and structure context.  }
  \label{vis_gopro}
   % Best viewed on a screen and zoomed in.
   \vspace*{-14pt}
\end{figure}
\vspace*{-12pt}

\section{Experiments}
\vspace{-6pt}
\subsection{Datasets}
\vspace{-6pt}
We evaluate our method on both synthetic and real-world event-based deblurring datasets.
Specifically, we adopt the widely used GoPro dataset~\cite{nah2017GoPro}, which provides pairs of blurry and sharp images synthesized by averaging high-speed frames. 
GoPro contains 3,214 image pairs at a resolution of $1280 \times 720$, including 2,103 pairs for training and 1,111 for testing.
To further assess the robustness, we leverage 412 blurry images of size $1632 \times 1224$ from the HighREV dataset~\cite{sun2023event}, along with their corresponding event streams. 
Besides, six sequences from the REVD dataset~\cite{kim2024frequency} are adopted, which includes 1,359 blurry images captured in typical urban scenes with diverse motion patterns, such as ego-motion, object motion, and their combinations.

\vspace*{-12pt}
\subsection{Implementation Details}
\vspace{-6pt}
In the training process, $256 \times 256$ patches are cropped, and 400k iterations are conducted during the training process.
Besides, Equation \ref{ur} is set as $\alpha \sim \mathcal{U}(\alpha_{\min} = 0, \alpha_{\max} = 0.2)$.
% data augmentation techniques such as random horizontal/vertical flips, rotations, and
We employ the Adam optimizer with an initial learning rate of $2 \times 10^{-4}$, $(\beta_1, \beta_2) = (0.9, 0.999)$ and $\epsilon = 10^{-8}$.
To maintain the training stability, a gradual warmup strategy in \cite{Zamir2021MPRNet} is applied in our training process with a minimum learning rate of $1 \times 10^{-6}$.
The evaluation metrics are the Peak Signal-to-Noise Ratio (PSNR) and the Structural Similarity Index Metric (SSIM).

\begin{table}[!t]
  \centering
  \caption{\textbf{Comparisons on GoPro dataset.} Metrics are PSNR ($\uparrow$) and SSIM ($\uparrow$). UR stands for the various under-reporting ratio on events. DSTN denotes image deblurring without event assistance. }
  \vspace*{-10pt}
  \label{mask_influence_all}%
  \renewcommand\arraystretch{1.5}
  \resizebox{\linewidth}{!}{
    \begin{tabular}{c|c|ccccc|c}
    \toprule[1.5pt]
    UR & DSTN & EFNet & STCNet & TRMD  & AHDINet   & MAT & Ours \\
    \hline
    0     & 35.05 / 0.973 & 35.47 / 0.972 & 36.46 / 0.975 & 36.58 / 0.979 & 37.09 / 0.978 & 36.67 / 0.978 & \textbf{37.63 / 0.9802} \\
    0.05  & 35.05 / 0.973 & 35.25 / 0.971 & 35.93 / 0.973 & 35.77 / 0.976 & 35.78 / 0.973 & 35.77 / 0.974 & \textbf{37.55 / 0.9800} \\
    0.1   & 35.05 / 0.973 & 34.92 / 0.970 & 35.39 / 0.972 & 35.05 / 0.973 & 34.77 / 0.969 & 35.53 / 0.973 & \textbf{37.45 / 0.9797} \\
    0.15  & 35.05 / 0.973 & 34.52 / 0.968 & 34.84 / 0.970 & 34.42 / 0.970 & 33.90 / 0.964 & 35.20 / 0.972 & \textbf{37.34 / 0.9794} \\
    0.2   & 35.05 / 0.973 & 34.08 / 0.966 & 34.29 / 0.967 & 33.84 / 0.967 & 33.15 / 0.960 & 34.81 / 0.970 & \textbf{37.21 / 0.9790} \\
    0.3   & 35.05 / 0.973 & 33.16 / 0.960 & 33.15 / 0.962 & 32.88 / 0.961 & 31.81 / 0.951 & 33.79 / 0.965 & \textbf{36.89 / 0.9781} \\
    \bottomrule[1.5pt]
    \end{tabular}%
  }
  \vspace{-12pt}
\end{table}%

\begin{figure}[!t]
  \centering
  \includegraphics[width=0.90\linewidth]{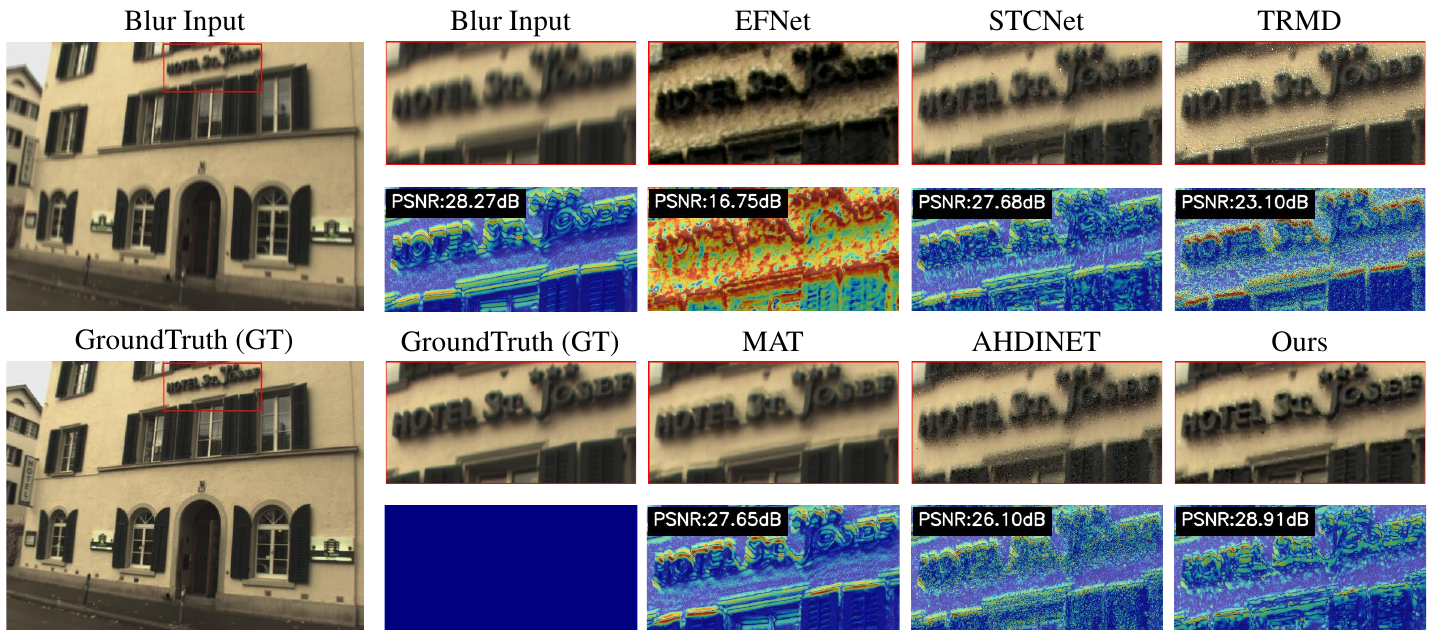}
  \vspace{-6pt}
  \caption{Visualization in HighREV dataset. }
  \label{vis_high}
   % Best viewed on a screen and zoomed in.
   \vspace*{-16pt}
\end{figure}

\vspace*{-12pt}
\subsection{Comparable Results}
\vspace{-6pt}
% methods chosen
Five state-of-the-art event-based image deblurring methods are chosen as fair and straight comparisons, including EFNet \cite{sun2022event}, STCNet \cite{yang2024motion},  TRMD\cite{chen2024motion}, AHDINet \cite{yang2025asymmetric}, and MAT \cite{xu2025motion}. Besides, DSTN \cite{Pan_2023_CVPR} as an image deblurring method, is adopted as a reference to verify the effectiveness of introducing events. 

\noindent{\textbf{Comparable Results in GoPro:}} 
Table~\ref{mask_influence_all} reports quantitative comparisons on the GoPro dataset.
Across different under-reporting ratios (UR), RED consistently achieves the highest PSNR and SSIM.
As shown in Figure~\ref{fig2}, existing event-based methods exhibit a sharp performance drop as UR increases, indicating their inability to extract effective motion priors from over-reporting events.
This finding echoes our motivation: simply injecting event inputs without considering modality-specific characteristics renders models highly vulnerable to event corruption.
In contrast, RED maintains stable performance even at UR = 0.5, outperforming the image-only baseline.
This robustness is largely attributed to the diverse under-reporting patterns introduced by our RPS module and our `distangle first and selective fusion' strategy.
Furthermore, visual results in Figure~\ref{vis_gopro} demonstrate that our method produces sharper details and cleaner textures.
% \textit{More detailed numerical results, qualitative results, and performance analysis are presented in \textbf{Appendix}}.

\begin{figure}[!t]
  \centering
  \includegraphics[width=0.95\linewidth]{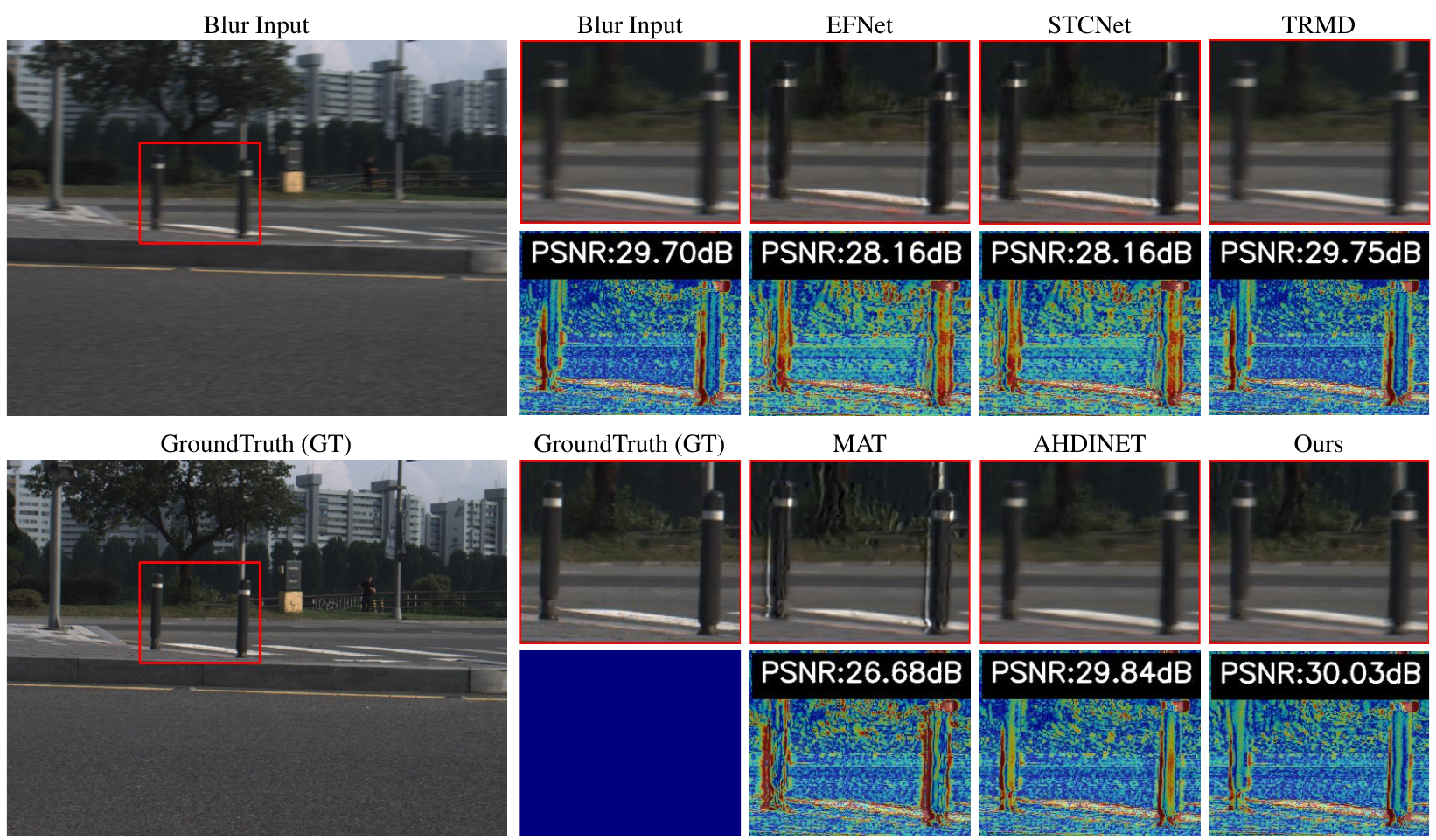}
  \vspace{-6pt}
  \caption{Visualization in REVD dataset.}
  \label{vis_red}
   % Best viewed on a screen and zoomed in.
   \vspace*{-18pt}
\end{figure}

\noindent{\textbf{Comparable Results in HighREV and REVD:}} 
To further verify the generalization in practice, we conduct deblurring experiments on HighREV and REVD datasets with the former trained models on GoPro.
As shown in Tables~\ref{com_results_highrev} and \ref{com_results_fevd}, our method consistently achieves the highest PSNR and SSIM, demonstrating strong robustness across diverse scenes and motion patterns.
Besides, visual comparisons on the HighREV dataset in  Figure \ref{vis_high} and the challenging REVD dataset in Figure \ref{vis_red} show Ours achieves clearer results.
Interestingly, we observe that PSNR of the deblurred image in some methods is lower than that of the blurry input.
This phenomenon indicates their failure to extract useful motion priors from real-world events for deblurring, further validating the motivation behind our approach.
\textit{More details are presented in} \textbf{\textit{Supplementary}}.

\begin{table}[!t]
  \centering
  \caption{\textbf{Quantitative comparisons.}  Metrics are PSNR ($\uparrow$) and SSIM ($\uparrow$). Best results on HighREV (\textbf{\emph{Left}}) and REVD (\textbf{\emph{Right}}) datasets are in \textbf{bold}.}
  \vspace*{-4pt}
  % ----- Left subtable: HighREV -----
  \begin{subtable}[t]{0.48\textwidth}
    \centering
    \renewcommand\arraystretch{1.4}
    \resizebox{\linewidth}{!}{
    \begin{tabular}{c|ccccc|c}
      \toprule[1.5pt]
      Methods & EFNet & STCNet & TRMD  & MAT   & AHDINet & Ours \\
      \midrule
      PSNR($\uparrow$)  & 20.56 & 28.77 & 28.89 & 27.54 & 28.82 & \textbf{30.04} \\
      SSIM($\uparrow$)  & 0.808 & 0.869 & 0.892 & 0.922 & 0.857 & \textbf{0.929} \\
      \bottomrule[1.5pt]
    \end{tabular}%
    }
    \caption{Comparisons on HighREV dataset.}
    \label{com_results_highrev}%
  \end{subtable}
  \hfill
  % ----- Right subtable: REVD -----
  \begin{subtable}[t]{0.48\textwidth}
    \centering
    \renewcommand\arraystretch{1.4}
    \resizebox{\linewidth}{!}{
    \begin{tabular}{c|ccccc|c}
      \toprule[1.5pt]
      Methods & EFNet  & STCNet & TRMD & MAT   & AHDINet & Ours \\
      \midrule
      PSNR($\uparrow$)  & 26.17 & 26.97 &  26.88 & 24.83     & 26.30 & \textbf{27.35} \\
      SSIM($\uparrow$)  & 0.829 & 0.857 &   0.858 & 0.803   & 0.822 & \textbf{0.860} \\
      \bottomrule[1.5pt]
    \end{tabular}%
    }
    \caption{Comparisons on REVD dataset.}
    \label{com_results_fevd}%
  \end{subtable}
  \label{tab:highrev_revd}
  \vspace{-18pt}
\end{table}

% \begin{figure}[t]
%     \centering
%     \vspace{-4pt}
%     \begin{minipage}[t]{0.56\textwidth}
%         \centering
%         \includegraphics[width=\linewidth]{./Fig/Result/vis_high1.pdf}
%         \caption{Visualization in HighREV.}
%         \label{vis_high}
%     \end{minipage}\hfill
%     \begin{minipage}[t]{0.42\textwidth}
%         \centering
%         \includegraphics[width=\linewidth]{./Fig/Result/vis_revd1.pdf}
%         \caption{ Visualization in REVD.}
%         \label{vis_red}
%     \end{minipage}
%     \vspace*{-18pt}
% \end{figure}

\vspace*{-12pt}
\subsection{Ablation Study}
\vspace*{-6pt}

\begin{figure}[!t]
  \centering
  % a
  \begin{subfigure}[t]{0.24\textwidth}
    \centering
    \captionsetup{skip=2pt}
    \includegraphics[width=\linewidth]{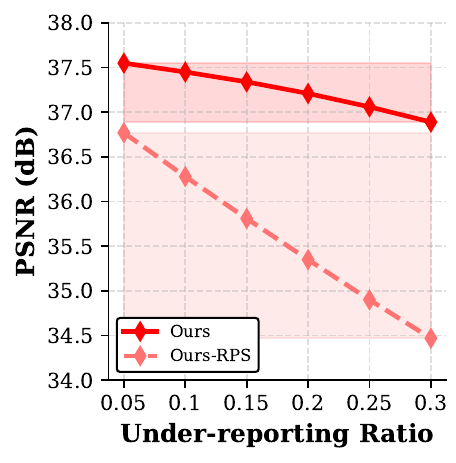}
    \caption{}
    \label{cmp_ours_rps}
  \end{subfigure}%
  \hfill
  % b
  \begin{subfigure}[t]{0.24\textwidth}
    \centering
    \captionsetup{skip=2pt}
    \includegraphics[width=\linewidth]{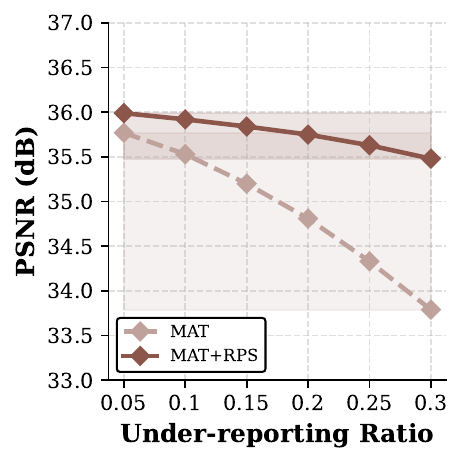}
    \caption{}
    \label{cmp_mat}
  \end{subfigure}%
  \hfill
  % c
  \begin{subfigure}[t]{0.24\textwidth}
    \centering
    \captionsetup{skip=2pt}
    \includegraphics[width=\linewidth]{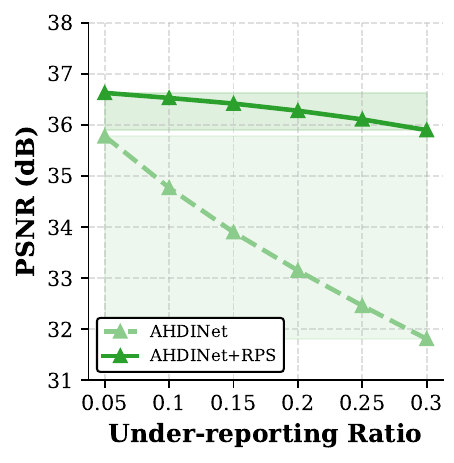}
    \caption{}
    \label{cmp_ahdinet}
  \end{subfigure}%
  \hfill
  % d
  \begin{subfigure}[t]{0.24\textwidth}
    \centering
    \captionsetup{skip=2pt}
    \includegraphics[width=\linewidth]{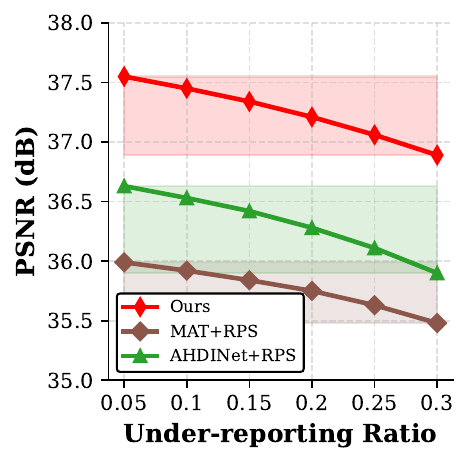}
    \caption{}
    \label{cmp_ours_all}
  \end{subfigure}
  \vspace*{-4pt}
  \caption{ (a) Ours v.s. Ours w/o RPS, (b) MAT v.s. MAT+RPS, (c) AHDINet v.s. AHDINet+RPS, and (d) Ours v.s. MAT+RPS/AHDINet+RPS. 
  } 
  \label{fig_ablation_rps}
  \vspace*{-20pt}
\end{figure}

\noindent{\textbf{Investigation of RPS.}}
To evaluate the effectiveness of our RPS, we first validate RED and RED without RPS under different UR settings in Figure~\ref{cmp_ours_rps}. 
Then, we integrate RPS into existing event-based methods to further assess its efficiency, including Transformer-based MAT~\cite{xu2025motion} in Figure~\ref{cmp_mat} and CNN-based AHDINet~\cite{yang2025asymmetric} in Figure~\ref{cmp_ahdinet}.
Finally, an overall comparison in Figure~\ref{cmp_ours_all} is conducted on Ours, MAT with our RPS, and AHDINet with our RPS.
From Figure~\ref{fig_ablation_rps}, the following conclusions are drawn. 
First, removing RPS from our framework leads to significant robustness drops. 
Second, incorporating RPS into existing event-guided methods consistently improves their robustness, demonstrating the plug-and-play generality of RPS.
Finally, beyond the contribution of RPS, our method consistently achieves the best performance and robustness across all conditions, highlighting our modality-specific distanglement.

To further verify the efficiency of our RPS, as formulated in Equ. \ref{bernolli} and \ref{ur}, the $\text{FLOPs} = TCHW + 3TCHW \cdot DR = TCHW (1 + 3DR)$,  where $T$ is the number of event frames, $(C, H, W)$ are the channel and spatial dimensions, and $DR$ denotes the disrupted ratio. To further quantify the overhead, we test the runtime of RPS over 100 runs on an event tensor of size $(B, T, C, H, W) = (1, 6, 1, 360, 640)$. RPS introduces only 2.49M additional FLOPs and roughly 0.71 ms of runtime overhead when applied to the above tensor.

\begin{table}[!t]
  \centering
\captionsetup{skip=4pt} 
  \caption{\textbf{Ablation study}. Metrics are PSNR ($\uparrow$) and SSIM ($\uparrow$). }
 \vspace{-6pt}
  \begin{subtable}[t]{0.50\textwidth}
    \centering
    \renewcommand\arraystretch{1.0}
    \resizebox{\linewidth}{!}{
    \begin{tabular}{ccccc}
      % \toprule[1.2pt]
      \hline
      Semantic & Motion & Cross & PSNR($\uparrow$) & SSIM($\uparrow$) \\
      \midrule
      \XSolidBrush & \XSolidBrush & \XSolidBrush & 25.77 & 0.864 \\
      \XSolidBrush & \Checkmark   & \Checkmark   & 34.99 & 0.964 \\
      \Checkmark   & \XSolidBrush & \Checkmark   & 35.04 & 0.966 \\
      \Checkmark   & \Checkmark   & \XSolidBrush & 37.14 & 0.978 \\
      \Checkmark   & \Checkmark   & \Checkmark   & \textbf{37.63} & \textbf{0.980} \\
      % \bottomrule[1.5pt]
      \hline
    \end{tabular}
    }
    \caption{Ablation study on MRM. }
    \label{abla_attn}
  \end{subtable}
  \hfill
  % ----- Right subtable: MSEM & ESEM -----
  \begin{subtable}[t]{0.45\textwidth}
    \centering
    \renewcommand\arraystretch{1.0}
    \resizebox{\linewidth}{!}{
    \begin{tabular}{cccc}
      % \toprule[1.2pt]
      \hline
      MSEM & ESEM & PSNR($\uparrow$) & SSIM($\uparrow$) \\
      \midrule
      \XSolidBrush & \XSolidBrush & 36.78 & 0.976 \\
      \XSolidBrush & \Checkmark   & 36.97 & 0.977 \\
      \Checkmark   & \XSolidBrush & 37.18 & 0.978 \\
      \Checkmark   & \Checkmark   &  \textbf{37.63} & \textbf{0.980} \\
      % \bottomrule[1.4pt]
      \hline
    \end{tabular}
    }
    \caption{Ablation study on MSEM and ESEM.}
    \label{abla_msem_ESEM}
  \end{subtable}
  \label{tab:abla_study}
  \vspace*{-20pt}
\end{table}

\noindent{\textbf{Investigation of MRM.}} 
To assess MRM, we conduct a comprehensive ablation study by selectively replacing the semantic reasoning, motion-wise structural representation, and cross-modality attention mechanisms with a modality-agnostic self-attention module~\cite{yuan2021tokens}, while keeping the feature dimensionality unchanged.
As shown in Table~\ref{abla_attn}, replacing all specialized attentions with generic self-attention causes a dramatic performance drop of 11.86 dB in PSNR, indicating that indiscriminately processing blurry images and events severely impairs effective feature extraction and interaction.
Replacing only the modality-specific attention in the image or event branch results in 2.64dB and 2.59dB drops, while a smaller but noticeable degradation of 0.49dB when replacing the cross-modality attention.
These findings collectively demonstrate that our MRM, not only captures the distinct semantic context and motion-sensitive areas, but also facilitates more effective and complementary feature fusion.

\noindent{\textbf{Investigation of MSEM and ESEM.}} 
We further investigate the contributions of two feature interaction modules in our RED.
As reported in Table~\ref{abla_msem_ESEM}, each module individually improves PSNR and SSIM over the baseline. 
The joint cooperation of MSEM and ESEM leads to the best performance, with a 0.85dB gain in PSNR over the baseline.
Besides, we observe that the performance gain from feature interaction strategies is less pronounced than that from modality-specific attention in Table~\ref{abla_attn}. 
This also supports our hypothesis: 
modality-specific feature encoding benefits modality-agnostic representation and enables more effective cross-modality collaboration, ultimately superior performance.

\vspace{-12pt}
\section{Conclusion}
\vspace{-9pt}
In this paper, we present RED, a robust and practical event-based deblurring framework that explicitly accounts for under-reporting events in practice.
We introduce RPS to embody RED with powerful and robust adaptability under different unknown scenario conditions.
To extract fine-grained motion priors from under-reporting events and prevent corrupted effects, a disentangled MRM is designed to explicitly model semantic, motion, and cross-modality correlations, allowing more effective extraction of modality-specific features and cross-modality fusion.
Building on these reliable features, MSEM and ESEM are performed to delineate blurry images with robust motion-sensitive priors and complement incomplete events with holistic semantic context. 
With the robust event representation and modality-specific disentangled representation, our RED delivers state-of-the-art deblurring performance and strong generalization.

\vspace{-12pt}

% \clearpage\mbox{}Page \thepage\ of the manuscript.
% \clearpage\mbox{}Page \thepage\ of the manuscript.
% \clearpage\mbox{}Page \thepage\ of the manuscript.
% \clearpage\mbox{}Page \thepage\ of the manuscript.
% \clearpage\mbox{}Page \thepage\ of the manuscript. This is the last page.
% Now we have reached the maximum length of an ECCV \ECCVyear{} submission (excluding references and acknowledgements).
% References should start immediately after the main text, but can continue past p.\ 14 if needed. 
% \clearpage  % TODO FINAL: This \clearpage needs to be removed from both review and camera-ready versions.

\bibliographystyle{splncs04}
\bibliography{main}
\end{document}